\documentclass[balance,upint,subscriptcorrection,varvw,mathalfa=cal=boondoxo,spanish,french,vietnamese,russian,greek,pdf-a,colorlinks,nofoot]{asmeconf}

\usepackage{makecell}
\usepackage{fancyhdr}
\usepackage{geometry}
\allowdisplaybreaks 

\hypersetup{%
	pdfauthor={Manuel Sage},									  
	pdftitle={Enhancing Battery Storage Energy Arbitrage with Deep Reinforcement Learning and Time-Series Forecasting},                  
	pdfkeywords={energy arbitrage, energy storage, reinforcement learning},
	pdfsubject = {Conference paper on energy arbitrage with batteries.},			  
}

\fancypagestyle{plain}{  
	\fancyhf{} 
	\fancyhead[L]{\textit{This paper is a preprint accepted for publication at ASME}} 
	\fancyfoot[C]{\thepage} 
}

\begin{document}





\title{Enhancing Battery Storage Energy Arbitrage With \\Deep Reinforcement Learning and Time-Series Forecasting}
 
%
%
%

\SetAuthors{%
	Manuel Sage\affil{1}, 
	Joshua Campbell\affil{1}, 
	Yaoyao Fiona Zhao\affil{1}\CorrespondingAuthor{yaoyao.zhao@mcgill.ca}, 
	}

\SetAffiliation{1}{McGill University, Montreal, Canada}

\maketitle





\begin{abstract}
Energy arbitrage is one of the most profitable sources of income for battery operators, generating revenues by buying and selling electricity at different prices. Forecasting these revenues is challenging due to the inherent uncertainty of electricity prices. Deep reinforcement learning (DRL) emerged in recent years as a promising tool, able to cope with uncertainty by training on large quantities of historical data. However, without access to future electricity prices, DRL agents can only react to the currently observed price and not learn to plan battery dispatch. Therefore, in this study, we combine DRL with time-series forecasting methods from deep learning to enhance the performance on energy arbitrage. We conduct a case study using price data from Alberta, Canada that is characterized by irregular price spikes and highly non-stationary. This data is challenging to forecast even when state-of-the-art deep learning models consisting of convolutional layers, recurrent layers, and attention modules are deployed. Our results show that energy arbitrage with DRL-enabled battery control still significantly benefits from these imperfect predictions, but only if predictors for several horizons are combined. Grouping multiple predictions for the next 24-hour window, accumulated rewards increased by 60\% for deep Q-networks (DQN) compared to the experiments without forecasts. We hypothesize that multiple predictors, despite their imperfections, convey useful information regarding the future development of electricity prices through a “majority vote” principle, enabling the DRL agent to learn more profitable control policies.
\end{abstract}







\section{Introduction}
\subsection{Background}
Energy arbitrage (EA) describes the practice of buying electricity when prices are low and selling it when prices are high. When battery energy storages (BESs) are involved, charging and discharging occur at low and high power prices, respectively \cite{a213_CAO_energyarbitrage, a241_wang_ea_with_rl}. The purpose of EA with BESs is to reduce costs or increase revenues for the battery operators. The growing incorporation of intermittent renewable energy into electricity grids comes with an increasing demand for flexible energy storages. At the same time, intermittent renewables can increase price volatility \cite{volatility2_RINTAMAKI2017270, volatility3_CIARRETA2020104749}. This favors EA with batteries, which has become the largest profit opportunity for BES operators according to Ref. \cite{a213_CAO_energyarbitrage}.  

Both planning new BES projects and controlling existing BESs are challenging due to the stochastic nature of electricity prices. A control strategy is required that dispatches the BES accordingly to maximize performance. Traditional tools for optimizing BES dispatch include programming methods such as mixed-integer linear programming \cite{x2_BIGGINS2022104234, x3_METZ201827}. These white-box approaches can compute optimal solutions over small time horizons, but quickly become intractable for larger problems and time spans. Besides, they also require access to the dynamics of the environment and future states \cite{a176_zhang2020, a171_MENG202113}. Thus, the obtained solutions represent the ideal case of perfect future knowledge and might not be achievable in practice.
Another popular category of optimizers for EA with batteries are heuristic methods, such as genetic algorithms and similar evolutionary approaches \cite{h1_9334858, h2_FENG2022105508}. Heuristic methods only require access to a sample model which facilitates the implementation of non-linearities such as battery degradation or charging efficiencies. However, heuristic methods suffer from poor convergence properties and sample inefficiency, which causes high compute cost \cite{a176_zhang2020, a93_zhang2019deep}. Like white-box approaches, applications of heuristic models are limited to shorter time periods, affecting their ability to handle uncertainty.

Reinforcement learning (RL) has recently gathered interest in the energy systems domain as an alternative. RL is a subset of machine learning characterized by an agent that learns by interacting with its environment. By training on large amounts of historical data, RL can learn control policies while considering uncertainty. Recent advances in deep reinforcement learning (DRL), which combines RL frameworks with deep learning, have led to more powerful, stable, and sample-efficient algorithms \cite{a79_mnih2015human, a121_schulmanPPO}. RL only requires a sample model and no future information about uncertain variables. However, in the case of EA, providing the agent with information about future electricity prices might be beneficial. In this study, we therefore evaluate the combination of two deep learning methods: deep reinforcement learning and time-series forecasting with deep neural networks.  

Our study centers around the possible performance gains of this combination for energy arbitrage when dealing with messy, real-world data. We conduct a case study in Alberta, Canada that is characterized by great electricity price fluctuations and the absence of cyclic patterns. Comparing different predictors and two DRL algorithms, namely deep Q-networks (DQN) and proximal policy optimization (PPO), we show that time-series forecasting can significantly boost RL performance despite high forecasting errors. Our results indicate that performance highly depends on the forecasted horizon and the number of forecasters available, with multiple forecasters leading to better performances. 

\subsection{Related Work}
Historically, statistical, non-machine learning models such as autoregressive integrated moving average (ARIMA) techniques have been widely used for time-series forecasting \cite{ j3_arima_price_forecasting_and_feature_selection_8665889, j7_microeconomic_AB_https://doi.org/10.1002/asmb.2681}. Reference \cite{j7_microeconomic_AB_https://doi.org/10.1002/asmb.2681} used statistical and microeconomic models to predict price spikes in the Alberta electricity market from 2002-2015.
However, in recent years deep learning methods have become popular. Neural networks with recurrent and convolutional architectures have demonstrated superiority over simpler statistical methods \cite{j4_lstm_pv8483960, j5_cnn_lstm_hybrid_comparison_DEHGHANI2023102119, j6_lstmcnnhybrid_implementation_9356582}.
For example, Ref. \cite{j8_lstm_successPENG20181301} showed that a properly tuned Long Short-Term Memory (LSTM) model, a type of recurrent network, performed best when predicting hour-ahead and day-ahead electricity prices in New South Wales compared to other time-series forecasting techniques. More recently, Ref. \cite{j9_attention_price_lstmMENG2022124212} used LSTM models combined with an attention mechanism to achieve a mean absolute error (MAE) reduction of 7.09\% when forecasting electricity prices in the Denmark compared to the standard LSTM implementation. Reference \cite{j10_attention_hybrid_PV_paperQU2021120996} proposed a Convolutional Neural Network (CNN) combined with attention-based LSTM layers to form a hybrid model. This architecture achieved a 2.20\% improvement in normalized root-mean-squared error (RMSE) when predicting photovoltaic power one hour into the future compared to a regular LSTM model.

Typically, for time-series forecasting of environmental and energy variables, many input time-series are used to provide the models with additional information. This is balanced with the amount of available data, the quality of the data, and the need to prevent dilution. For example, Ref. \cite{j3_arima_price_forecasting_and_feature_selection_8665889} performed extensive feature selection for predicting day-ahead electricity price and concluded that hourly price, demand, wind power generation, wind speed, and ambient temperature were the most important factors for their case study of the Iberian electricity market and gave their models the best performance. Real-world electricity price data can be irregular which makes time-series forecasting more inherently difficult. Reference \cite{j11_smoothing_livieris_smoothing_2021} showed that smoothing the data as a preprocessing step can improve model performance when forecasting variables in the energy sector.

In the related work applying RL to EA with batteries, EA is either regarded as the sole task of the battery \cite{a213_CAO_energyarbitrage, a241_wang_ea_with_rl, a242_han_rl_and_dl_for_ea}, or combined with other battery services such as frequency regulation, demand response, load following, and improved utilization of renewable energies \cite{a225_huang, a237_HARROLD2022121958, a193_DASILVAANDRE2022108551, a226_DONG2021107229}. Reference \cite{a241_wang_ea_with_rl} conducted a simple study in which tabular Q-learning is applied to pure EA. Here, the RL agent makes decisions solely based on the current electricity price and the state of charge (SOC) of the battery. A similar approach, also with tabular Q-learning, was studied in Ref. \cite{a242_han_rl_and_dl_for_ea}. 

Reference \cite{a226_DONG2021107229} compared tabular Q-learning, Q-learning with linear function approximation, and Sarsa to particle swarm optimization on a combined frequency regulation and EA task. In the conducted case study, Q-learning with function approximation performed best. Reference \cite{a193_DASILVAANDRE2022108551} applied deep deterministic policy gradients (DDPG) to a joint load following and EA environment, where EA helps to reduce the cost of power supply. DDPG scored only slightly worse compared to an oracle based on MPC with access to perfect future information, but only required a fraction of the compute cost. Reference \cite{a225_huang} compared several DRL models including PPO, DDPG, and double DQN on a dispatch task combining frequency regulation, improved renewable energy utilization, and EA. This comparison is interesting as a value-based RL algorithm with discrete action choices (double DQN) is compared to three actor-critic methods with continuous actions. In the experiments, PPO performed best regarding both reward maximization and sample efficiency. 

In a few studies, RL has been deployed along with forecasts of uncertain variables to improve the performance on EA. Reference \cite{a242_han_rl_and_dl_for_ea} used LSTM cells to forecast electricity prices and demands for the next hour, which were then used in an EA environment controlled by tabular Q-learning. Reference \cite{a213_CAO_energyarbitrage} predicted the next 24 hours of electricity prices using a CNN-LSTM hybrid given the last 168 hours. The forecasts were then added to the state space of a noisy-net DQN model on a pure EA task. Reference \cite{a237_HARROLD2022121958} compared various variants of DQN to DDPG on a task combining EA, load following, and renewable energy control. The authors used artificial neural networks to forecast the next hour’s values for solar and wind power generation, demand, and electricity price. On the best performing model, rainbow DQN, these forecasts helped to increase rewards by 14\%.

\subsection{Contributions}

Our study differs from the existing work by its explicit focus on the possible performance gains on EA when combining time-series forecasting and DRL. We carefully benchmark different predictors, RL models, and forecasting horizons. Unlike the reviewed work, we conduct a case study on challenging electricity price data that is non-stationary and lacks obvious cyclic behavior. The data in Ref. \cite{a213_CAO_energyarbitrage, a242_han_rl_and_dl_for_ea, a237_HARROLD2022121958} shows clear patterns facilitating predictions. For example, the next hour price forecaster in Ref. \cite{a237_HARROLD2022121958} achieved a low mean absolute percentage error (MAPE) of 11.6\%, compared to 28.2\% in our case study (see section \ref{sec:forecast_results}). The contributions of this study can be summarized as follows:
\begin{enumerate}
    \item We formulate an EA task for a grid-connected BESs considering charge and discharge efficiencies as well as battery degradation. To this environment, we apply a combined deep learning framework consisting of DRL for battery control and time-series forecasting for predictions on future electricity prices.
    \item Conducting a case study on challenging price data from Alberta, Canada, we investigate how – despite high forecasting errors – DRL models can improve decision making when provided with price predictions. For this purpose, we benchmark different forecasting architectures, DRL models, and forecasting horizons. 
\end{enumerate}



\section{Problem Formulation} \label{sec:2_problem_form}

In a recent study currently under review, we have compared the performance of various DRL algorithms and experiment design choices on the same system \cite{sage2024}. In the present study, we extend this methodology with time-series forecasts and assess the effect on DRL performance.

\subsection{System Description}

We define a simple environment of a grid connected BES aiming to capitalize on the volatility of electricity prices. This setup, where performance largely depends on electricity prices, is most suitable to identify the influence of time-series forecasting on RL performance. We assume that the battery is a price-taker that can purchase and sell electricity at market prices without influencing prices. The objective is to maximize the revenues of energy arbitrage over the optimization period:

\begin{equation} \label{eq1}
	\max R_{Total} = \sum_{t=0}^{T} R_{grid, t} - C_{degr, t}
\end{equation}
\textit{subject to}
\begin{equation} \label{eq2}
	R_{grid, t} = c_{w, t} \times P_{B, t} \times \Delta t
\end{equation}
\begin{equation} \label{eq3}
	SOC_{t} = SOC_{t-1}(1-\sigma) - \eta \frac{P_{B, t}\Delta t}{C_{max}}
\end{equation}
\begin{equation} \label{eq4}
	C_{degr, t} = \frac{|(1 - SOC_t)^{k_p} - (1 - SOC_{t-1})^{k_p}|}{2 \times N^{fail}_{100}} \times C_{inv}
\end{equation}
\begin{equation} \label{eq5}
	P^{min}_{B} \leq P_{B,t} \leq P^{max}_{B}
\end{equation}
\begin{equation} \label{eq6}
	SOC^{min} \leq SOC_{t} \leq SOC^{max}
\end{equation}
\textit{where}
\begin{align*}  
t \hspace{0.1em}&= \text{time index}\\  
R_{grid} \hspace{0.1em}&= \text{revenues/cost from grid interaction}\\  
C_{degr} \hspace{0.1em}&= \text{cost of BES degradation}\\  
c_{w} \hspace{0.1em}&= \text{wholesale price of electricity}\\  
P_{B} \hspace{0.1em}&= \text{BES charging ($P_{B}<0$) / discharging power ($P_{B}>0$)}\\  
SOC \hspace{0.1em}&= \text{BES state of charge}\\  
\sigma \hspace{0.1em}&= \text{BES self-discharge}\\  
\eta \hspace{0.1em}&= \text{BES charge and discharge efficiency}\\  
C_{max} \hspace{0.1em}&= \text{BES capacity}\\  
k_p \hspace{0.1em}&= \text{Peukert constant}\\  
N^{fail}_{100} \hspace{0.1em}&= \text{number of full BES cycles until failure}\\
C_{inv} \hspace{0.1em}&=  \text{investment cost of BES}\\
P^{min}_{B} \hspace{0.1em}&= \text{BES charge limit}\\
P^{max}_{B} \hspace{0.1em}&= \text{BES discharge limit}\\
SOC^{min} \hspace{0.1em}&= \text{minimum allowable SOC}\\
SOC^{max} \hspace{0.1em}&= \text{maximum allowable SOC}\\
\end{align*}  

The total revenue at each time-step is composed of the revenue for grid interaction and the cost of battery degradation (Eq. \ref{eq1}). The grid revenue is computed with the current time-step’s electricity price and charging or discharging power of the battery (Eq. \ref{eq2}). Both grid revenue and battery power are positive when the battery is being discharged and negative when it is charged. We model the BES as a black box whose SOC changes depending on the applied charging or discharging power (Eq. \ref{eq3}). $\eta$ is the charging or discharging efficiency and $\eta = \eta_{dc} > 1$ in the case of discharging and $\eta = \eta_{ch} < 1$ in the case of charging. To model battery degradation (Eq. \ref{eq4}), we resort to a depth of discharge approach for cyclic battery ageing as presented in Ref. \cite{a226_DONG2021107229}. Equations (\ref{eq5}) and (\ref{eq6}) are constraints limiting the charging/discharging power of the BES and the SOC, respectively.

\subsection{Markov Decision Process}

To apply RL to this environment, we formulate a Markov Decision Process (MDP) consisting of state space $\mathcal{S}$, action space $\mathcal{A}$, transition function $\mathcal{P}$, reward function $\mathcal{R}$, and discount factor $\gamma$. Together, these elements form the tuple $(\mathcal{S}, \mathcal{A}, \mathcal{P}, \mathcal{R}, \gamma)$ \cite{a66_sutton2018reinforcement}.

\subsubsection{State Space}
A state $s_t \in \mathcal{S}$ contains the information available to the RL agent at the current time step. After taking an action, the environment transition to the next state $s_{t+1} \in \mathcal{S}$ according to the transition function $\mathcal{P}$.
In the basic experiments, i.e. without price forecasts, each state consists of the current battery SOC and electricity price:
\begin{equation} \label{eq7}
	s_t = (SOC_t, c_{w,t})
\end{equation}
In the experiments with price forecasts, one or more predictions, denoted $p$, are added to the state space:
\begin{equation} \label{eq8}
	s_t = (SOC_t, c_{w,t}, p_{t+1}, ..., p_{t+24})
\end{equation}

\subsubsection{Action Space} \label{sec:action_space}
The action space for the EA task is one dimensional and continuous. It reaches from the BES charge limit to the BES discharge limit: $a_t \in [P^{min}_{B}, P^{max}_{B}]$.
To avoid the violation of constraints (\ref{eq5}) and (\ref{eq6}), we implement a hard-coded safety layer that corrects the RL agent's actions if necessary:
\begin{equation} \label{eq9}
    a_{c,t} = 
    \begin{cases}
        \min(a_t, P^{max}_{B}, \frac{(SOC_{t}-SOC^{min})C_{max}}{\Delta t}) & , a_t \geq 0\\
        \max(a_t, P^{min}_{B}, \frac{(SOC_{t}-SOC^{max})C_{max}}{\Delta t}) & , a_t < 0
    \end{cases}
\end{equation}
where $a_{c,t}$ is the corrected action. For example, the safety layer prevents the agent from discharging the battery below $SOC^{min}$. We have used a similar mechanism in our previous work \cite{sage2023battery}.

\subsubsection{Reward Function}
The instantaneous reward at each time step ($R_t \in \mathcal{R}$) is $R_t = R_{grid, t} - C_{degr, t}$ (see Eq. \ref{eq1}). The RL agent seeks to maximize the return, which is the sum of discounted rewards:
\begin{equation} \label{eq10}
	G_t = R_{t+1} + \gamma R_{t+2} + \gamma^2 R_{t+3} + ... = \sum_{k=t+1}^{T}\gamma^{k-t-1} R_{k}
\end{equation}
where $\gamma$ is the discount factor and $\gamma \in [0,1]$ \cite{a66_sutton2018reinforcement}. We treat $\gamma$ as a hyperparameter that we tune in our experiments.

\section{Methodology} \label{sec:3_method}

\subsection{Reinforcement Learning}

In this section we focus on the description of the DQN algorithm \cite{a79_mnih2015human}, which performed best in our experiments. DQN and other value-iteration methods such as Q-learning and SARSA have dominated the reviewed related work and are popular in RL applications to battery dispatch in general \cite{a194_subramanya9777914}.
Value-iteration methods iteratively learn state values, $V(s)$, that quantify the value of being in a state, or state-action values, $Q(s, a)$, that quantify the value of being in a state and taking an action. In Q-learning, the Q-values are updated at each time-step using the Bellman equation:
\begin{equation} \label{eq11}
    Q(s,a) =  Q(s,a) + \alpha  [r + \gamma \max_{a'} (Q(s',a')) - Q(s,a)]
\end{equation}
where $\alpha$ is the learning rate and $r$ the immediate reward. $s'$ and $a'$ denote the next state and next action, respectively. This recursive equation allows the agent to update a Q-value based on the maximum possible Q-value in the next state \cite{a66_sutton2018reinforcement}.

In DQN, the Q-values are learned using neural networks as function approximators and denoted as $Q(s,a; \theta)$, where $\theta$ are the weights of the neural network. The key concepts of DQN are:
\begin{itemize}
    \item \textbf{Epsilon-greedy exploration:} The tradeoff between exploration and exploitation in DQN is managed through the $\epsilon$-greedy strategy:
    \begin{equation} \label{eq12}
        a = 
        \begin{cases}
            \arg \max_a Q(s,a;\theta), & \text{with probability } 1-\epsilon\\
            \text{random action}, & \text{with probability } \epsilon
        \end{cases}
    \end{equation}
    $\epsilon$ is often chosen to encourage exploration early during training and then annealed to favor exploitation. In our experiments, we treat $\epsilon$ and its behavior during training as hyperparameter that we tune to maximize performance.
    \item \textbf{Experience replay:} DQN stores the experiences made by interacting with the environment in the form of (\textit{state, action, reward, next state})-tuples in a replay buffer. At training time, a batch of interactions is then sampled from the buffer to compute the update. By decorrelating experiences, this strategy improves stability. At the same time, using a single experience for multiple updates improves sample efficiency. 
    \item \textbf{Target network:} DQN keeps a copy of the Q-network, called target network, that is updated periodically with the weights of the Q-network. The target network is used to compute the maximum possible next state-action value in the loss function:
    \begin{equation} \label{eq13}
        L(\theta) = \mathbb{E}_{s,a,r,s'}[(r+\gamma \max_{a'} Q(s',a';\theta^-) - Q(s,a;\theta))^2]
    \end{equation}
    where $\theta^-$ are the parameters of the target network. The loss is then differentiated with respect to the weights and used for gradient descent. The use of a target network has shown to reduce the overestimation bias and increase stability.    
\end{itemize}
A detailed explanation of these concepts and pseudocode of the DQN algorithm can be found in Ref. \cite{a79_mnih2015human}. Besides DQN, we also run experiments using PPO, an on-policy, actor-critic algorithm. For details on this model, the reader is referred to Ref. \cite{a121_schulmanPPO}.
DQN requires the discretization of the action space. After conducting preliminary experiments, we decide to divide the action space into three discrete actions: $a_t \in [P^{min}_{B}, 0, P^{max}_{B}]$, representing maximum charge, idle, and maximum discharge. PPO can handle continuous action spaces and is applied to the unmodified action space introduced in section \ref{sec:action_space}.

\begin{figure*}
	\begin{center}
		\includegraphics[width=\textwidth]{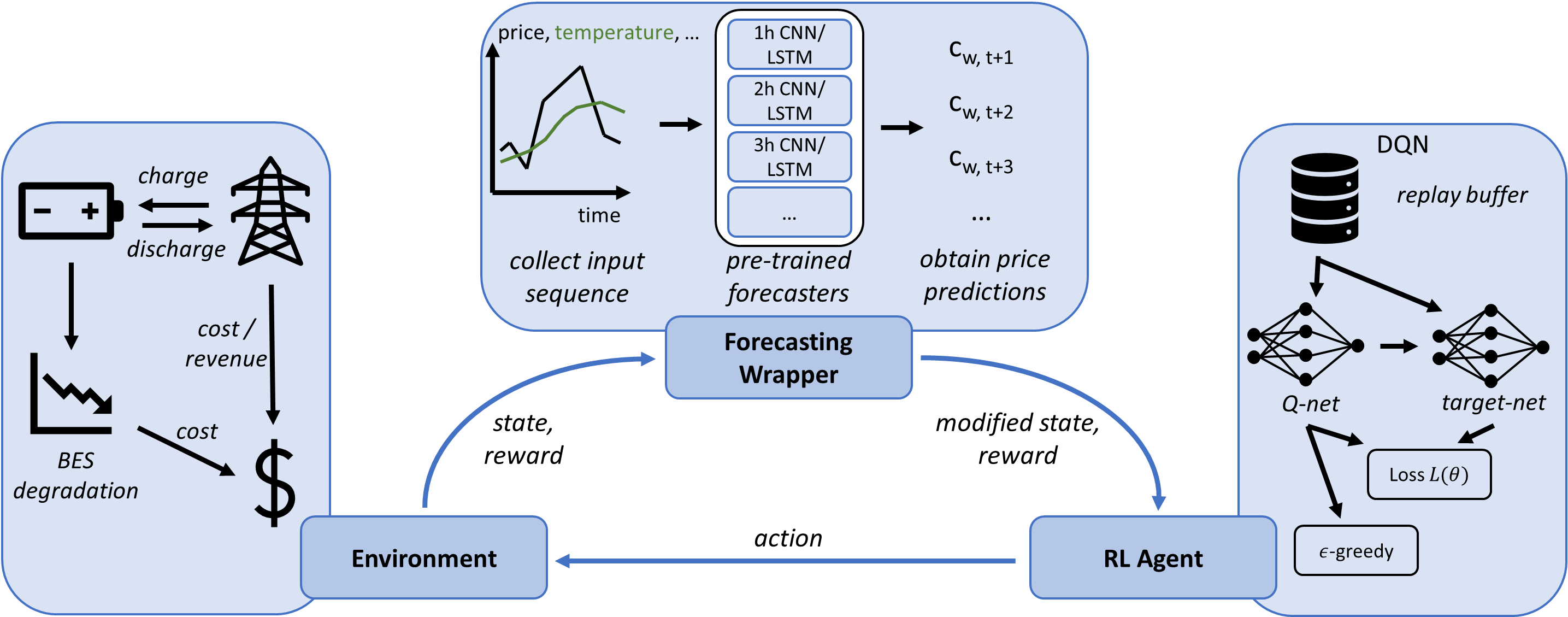}    
		\caption{Schematic showing the integration of time-series forecasters into the agent-environment framework of     reinforcement learning.} 
		\label{fig:method}
	\end{center}
\end{figure*}

\subsection{Time-Series Forecasting and Forecast Integration}

To predict electricity prices, we experiment with four different types of deep learning architectures that have demonstrated superior performance in the reviewed related work:
\begin{itemize}
    \item CNN \cite{original_CNN}
    \item LSTM \cite{j12_Original_LSTM}
    \item CNN-LSTM hybrids, with convolutional layer(s) being followed by LSTM-layer(s).
    \item CNN-LSTM hybrids with attention modules \cite{vaswani2017attention} after the LSTM layer(s).
\end{itemize}
All four model types end with one or more fully connected layers. The exact architectures, most prominently model depth, layer width, and activation function, are tuned to maximize performance. For convolutional layers, we tune kernel size and stride. Since we focus on point forecasts, each model is trained to forecast a single future electricity price and has thus one node in the output layer. This increases the number of models that must be tuned and trained compared to training one model to predict every horizon simultaneously, but it guarantees that the best performing model  can be obtained for each horizon.

The electricity data in the conducted case study has an hourly resolution. Based on the reviewed related work and preliminary experiments, we decide to train the forecasters to predict electricity prices in 1h, 2h, 3h, 6h, 12h, 18h, and 24h. These forecast horizons were identified to be the most useful for the RL agents (based on experiments supplying the agents with the future ground truth as the forecasts), and they require less computational resources than training every horizon 1-24h. For each of these seven horizons and four model types we conduct a separate hyperparameter tuning. The tuning includes experiments with different input features and window sizes. Besides the electricity price itself, we experiment with adding the electricity demand, ambient temperature, solar irradiance, wind speed, ambient pressure, and relative humidity. We further test for time-related features such as the hour of the day, the week of the year and the month of the year. The sequential data is processed into feature-label pairs using a "sliding window" approach, where the feature array is the combination of the input features for a sequence of data points in the past up to the current point and the label is the future electricity price for the horizon being trained. The length of the feature array is thereby the window size.

As a data preprocessing technique, we normalize all inputs using min-max scaling to a [-1, 1] range. We also experiment with smoothing the spiky and irregular price data, by converting the raw data to an exponentially weighted moving average of itself, as done in \cite{j11_smoothing_livieris_smoothing_2021}. In addition to the hyperparameters mentioned above, we tune batch sizes, learning rates and learning rate schedulers. We use early stopping as a regularization technique, Adam as optimizer, and the RMSE as loss function.

The integration of price forecasts into the RL framework is visualized in Fig. \ref{fig:method}. We use a training and validation set composed of data recorded prior to the optimization period of the RL agent. This ensures that the price data encountered by the RL agent was not seen by the forecasters beforehand.  Once the best forecasting models over the seven horizons are identified, these are saved and loaded into the forecasting wrapper. The wrapper sits between the environment and the agent, tracks the progress of the environment, and processes the input data for each forecaster. Then, the forecasters make predictions which are added to the original state and passed to the RL agent. Ideally, the RL agent makes better informed decisions with this additional information.


\section{Case Study} \label{sec:4_case_study}

\begin{figure*}
	\begin{center}
		\includegraphics[width=\textwidth]{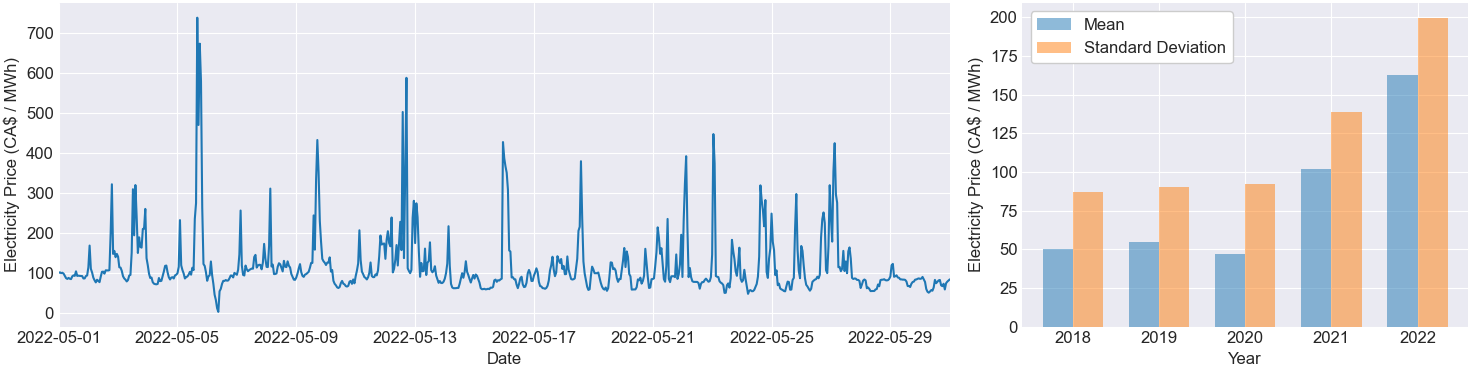}    
		\caption{(Left) A one month sample of electricity prices in Alberta from May 2022. (Right) Mean and       standard deviation of electricity prices in Alberta for the five years of data collected.}
		\label{fig:e-prices}
	\end{center}
\end{figure*}

\begin{table}
	\begin{center}
		\caption{Parameters of the case study}
		\label{tbl:params}
		\begin{tabular}{l|l}
			\textbf{Parameter}  & \textbf{Value}  \\ \hline
			battery capacity, $C_{max}$ & 10 MWh \\
			    $SOC^{min}$/ $SOC^{max}$ & 0.2 / 0.8  \\
			$P^{min}_{B}$ / $P^{max}_{B}$ & -2.5 / 2.5 MW \\
			charge/ discharge eff., $\eta$ & 0.92 / $\frac{1}{0.92}$\\
			self-discharge, $\sigma$ & 0 \\
			Peukert constant, $k_p$ & 1.14 \cite{a229_tran_bes}\\
			cycles to failure, $N^{fail}_{100}$ & 6,000 \cite{a230_cheng_bes}\\
			investment cost, $C_{inv}$ & 300,000 \$/MWh \cite{w23_cole_cost}\\
		\end{tabular}
	\end{center}
\end{table}

The location for the case study is in Alberta Canada, where the BES is connected to the utility grid and participating in price arbitrage. The parameters characterizing the BES are listed in Table \ref{tbl:params} and assume the operation of a Lithium-Ion battery. We obtain electricity data from the Alberta Electric System Operator \cite{aeso} and climate data from the ERA5 reanalysis model \cite{era5}. Both data types are downloaded in hourly resolution for the years 2018 to 2022. The first 3.5 years from January 2018 to June 2021 form the training set for the price forecasting models. The following six months between July 2021 and December 2022 are the validation set that we use to tune hyperparameters and identify the best forecasters. The year of 2022 serves as test set for the forecasters and optimization time frame for the RL models. For RL, we define the entire year, i.e. 8760 hours, as one episode. Figure \ref{fig:e-prices} provides additional information on the electricity data. The left part of the figure shows the challenging nature of electricity prices in Alberta, taking May 2022 as an example. Price spikes frequently occur, but at different times and with different magnitudes. The bar chart on the right highlights another challenge: the price data is highly non-stationary. Mean and standard deviation of electricity prices were relatively constant from 2018 to 2020. These three years form the majority of the training set. 2021 and especially 2022 show higher mean prices and more volatility, aggravating the forecasting task significantly. Our experiments therefore aim to tackle the challenge of improving energy arbitrage performance with price forecasts despite these challenges.

We utilize a variety of benchmarks for a fair assessment of the tested models. For time-series forecasting, we tune and train an ARIMA model. Additionally, a persistence model, naïvely assuming the forecasted electricity price is identical to the current price, serves as lower bound. Our lower bound for RL is the cross-entropy method (CEM), an evolutionary algorithm popular for continuous control tasks \cite{cem}. As upper bound, a genetic algorithm (GA) with access to environment dynamics and perfect future knowledge is used. In a model-predictive control (MPC) framework, the GA optimizes the sequence of battery control actions over a fixed horizon. Then, the first action of the obtained sequence is executed in the environment and the horizon shifts by one time-step in the future. This iterative process with receding horizon is repeated until the end of the episode is reached. Due to the full access to the environment dynamics, the MPC-GA model can be seen as an oracle whose performance is only limited by the available compute resources.

The codebase for this case study was written in Python 3.12 and is available along with the dataset at GitHub\footnote{\url{https://github.com/masa2203/battery_arbitrage_with_drl}}. PyTorch  was used as deep learning library for the time-series forecasters. Gymnasium and stable-baselines3 were used for the RL environment and agents, respectively. Both forecasters and DRL models were tuned using Optuna.


\section{Results and Discussion} \label{sec:5_results}

\begin{figure}
	\begin{center}
		\includegraphics[width=\linewidth]{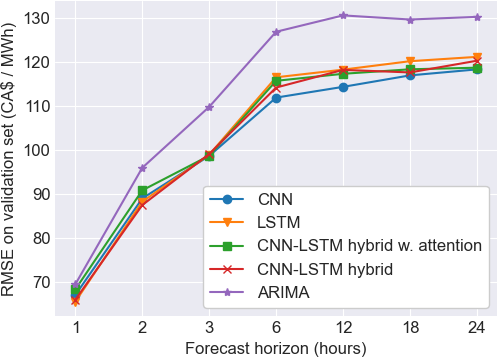}    
		\caption{RMSE of tested models on the validation set for different forecast horizons.}
		\label{fig:forecast_perf}
	\end{center}
\end{figure}

\subsection{Forecasting of Electricity Prices} \label{sec:forecast_results}

\begin{figure*}[hbt!]
	\begin{center}
		\includegraphics[width=\textwidth]{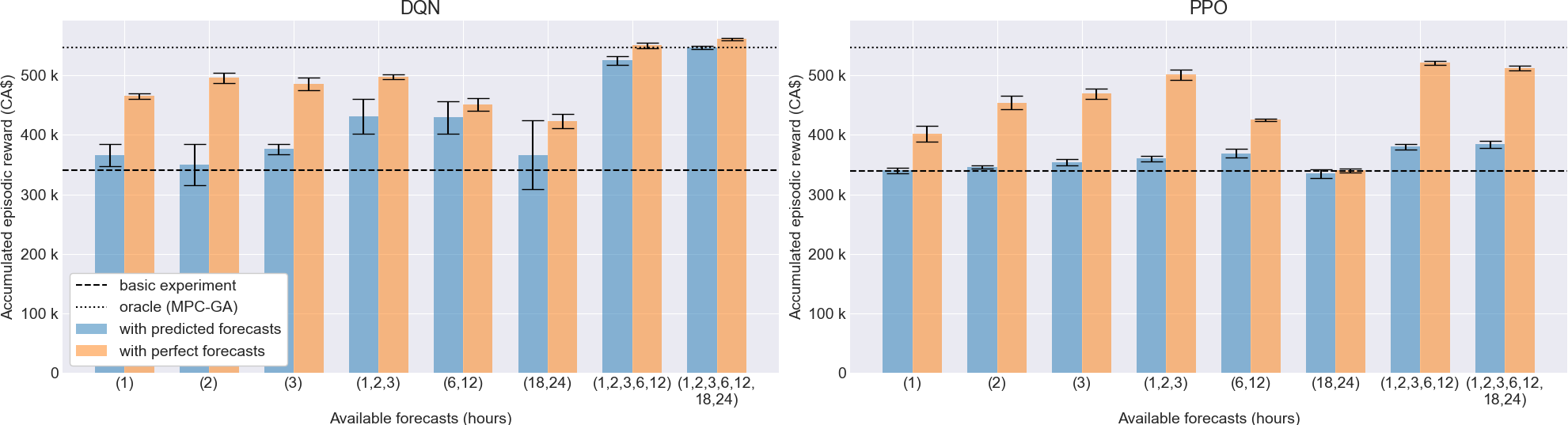}    
		\caption{DQN (left) and PPO (right) performance with different forecasting horizons for perfect (ground truth) and predicted forecasts after 50 episodes of training. All results were averaged over five independent runs. The error bars show $\pm$ one standard deviation.}
		\label{fig:main_perf}
	\end{center}
\end{figure*}

In Fig. \ref{fig:forecast_perf} we compare the performance on the validation set of the four deep learning models and ARIMA over different horizons after tuning. The persistence model scored significantly worse and was omitted in the figure. The forecasting error for all models increased with increasing forecast horizons. The four deep learning models performed similarly on all horizons, whereas the ARIMA model received higher errors. For 3h to 24h forecasts, the CNN was found to perform best. For 1h and 2h forecasts, the LSTM and CNN-LSTM hybrid performed best by small margins, respectively. We chose these models for the further experiments with RL. When scored on the test set, their performance metrics ranked from 100 – 175 for RMSE and 28\% - 68\% for MAPE. These errors are significant when put into perspective with the electricity price statistics in Fig. \ref{fig:e-prices} and the errors reported in the related work \cite{a237_HARROLD2022121958, a242_han_rl_and_dl_for_ea}. A sample of the predictive performance for selected horizons is provided in the top of Fig. \ref{fig:sample}. All three horizons plotted struggle with price spikes. 1h predictions lag behind spikes, while 12h and 24h predictions often miss spikes or predict spikes where they don’t occur and tend to estimate the magnitude of the spikes incorrectly.

We further noticed that simple architectures, with small and few layers, performed better than more complex structures. The best performing CNNs were all characterized by small kernel sizes (1-3) and strides (1-2), bringing them closer to regular, fully connected layers. Recurrent cells, the attention mechanism, and more complex convolutions all did not improve or even worsened results. Our interpretation of this is that the spiky and irregular price data caused higher-capacity models to learn patterns that were coincidental and non-predictive. The non-stationarity of the data, especially when comparing training, validation, and test sets, further aggravated the task, slightly favoring smaller models with less tendency to overfit.

Experiments with smoothened input data did not improve performance for any of the smoothing coefficients and horizons tested. Regarding input variables, model performance improved when adding the electricity demand of Alberta and hour of the day. The hour of the day was thereby encoded using a sine and a cosine wave to better convey the cyclic nature of the data \cite{sincos2_app12199788}.

\subsection{Energy Arbitrage with Price Forecasts}

We begin the analysis of the results of DRL on EA by looking at the basic experiments (no forecasts) and those with perfect forecasts (ground truth is provided). In the basic experiments (see dashed lines in Fig. \ref{fig:main_perf}), DQN and PPO score nearly identical and accumulate around CA\$340,000 in rewards throughout the episode. For the experiments with forecasts, we group the horizons into three ranges: short-term (1,2,3h), middle term (6,12h) and long-term (18,24h). When provided with perfect forecasts (orange bars in Fig. \ref{fig:main_perf}), both DQN and PPO profit the most from short-term forecasts. We therefore subdivided the short-term forecast into its components and found that DQN profited the most from perfect 2h forecasts, and PPO from perfect 3h forecasts. While still enabling performance gains, the 1h forecast was the least useful short-term forecast. This finding is interesting as some of the related work solely relied on forecasting the next hour, possibly missing out further improvements.
Our observation is underlined by the experiments combining short and middle-term forecasts and those with all forecasts. DQN and PPO further improved, with PPO scoring best with 1h, 2h, 3h, 6h, and 12h perfect forecasts and DQN with all seven horizons. In these two experiments, DQN outperformed even the oracle (MPC-GA). Comparing DQN and PPO on the experiments with perfect forecasts in Fig. \ref{fig:main_perf}, DQN achieved higher rewards on all combinations except for the short-term range. For both models, the stronger performances with 2h, 3h, and combined forecasts indicate that the agent requires time to react to changing prices. This is barely possible if only the next hour price is forecasted.

\begin{figure*}[hbt!]
	\begin{center}
		\includegraphics[width=\textwidth]{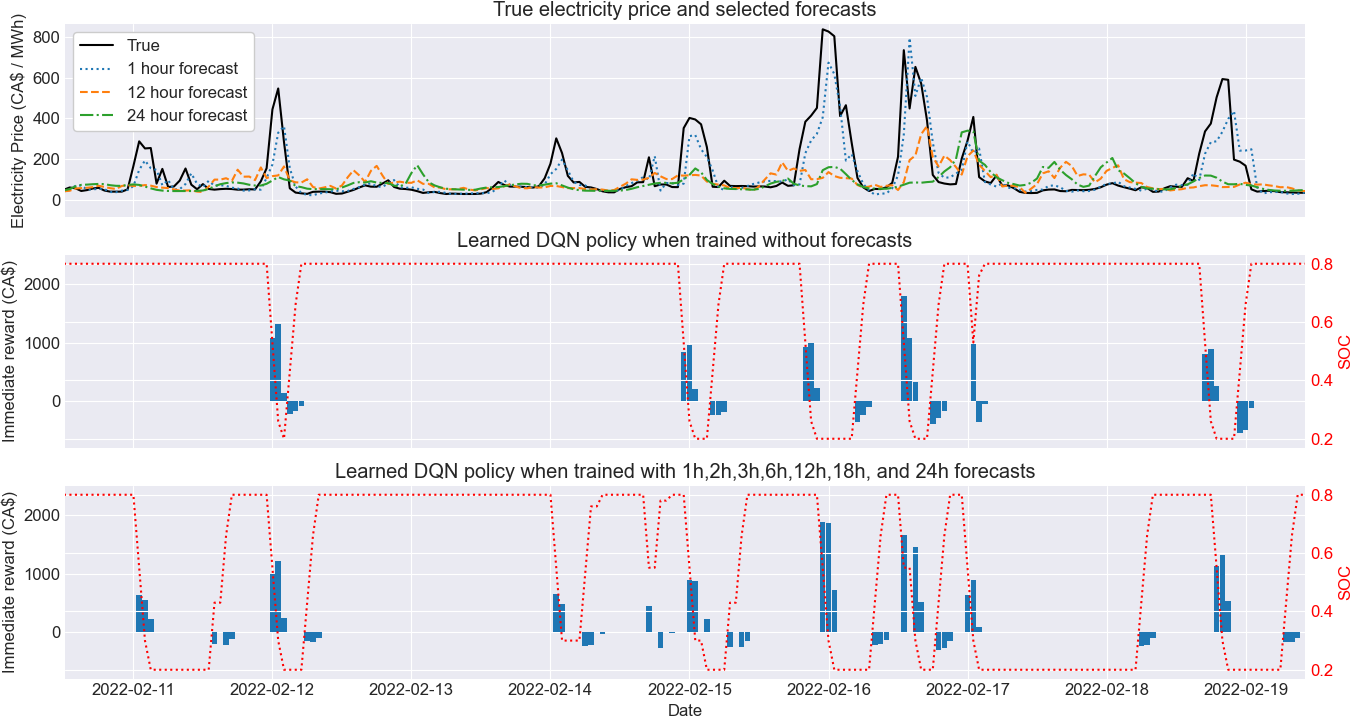}    
		\caption{A nine day sample of (Top) true electricity prices and predictions for selected horizons with the best deep learning models found, (Middle) the battery SOC and rewards for a trained DQN agent without access to forecasts, (Bottom) for a trained DQN agent with access to all seven predictions.} 
		\label{fig:sample}
	\end{center}
\end{figure*}

As a next step, we analyze the results of PPO and DQN with predicted forecasts (blue bars in Fig. \ref{fig:main_perf}). Naturally, the high prediction errors caused these experiments to score worse compared to perfect forecasts. For PPO, only moderate improvements compared to the basic experiment were achieved. Short-term and middle-term forecasts increased rewards by 6\% and 9\%, respectively. Long-term forecasts slightly decreased rewards. The best experiment, a combination of all seven forecasts, yielded an improvement of 14\%. DQN overall responded better to predicted forecasts but showed increased instability across independent runs for single forecasts and ranges. However, the combination of multiple forecast ranges resulted in high and stable rewards. These experiments also visibly narrowed the gap between predicted and perfect forecasts. With access to all seven forecasters, DQN reached an accumulated reward of CA\$547,000 or 60\% more than without forecasters (basic experiment). This represents a remarkable increase when considering the high forecasting errors.

To interpret these results, we compare the learned policies of basic DQN and DQN with access to all predictions. Figure \ref{fig:sample} shows a nine-day sample from both policies below true and predicted electricity prices for selected horizons during the same time interval. A striking difference between both policies is the increased activity for the agent with access to forecasts. Throughout the entire one-year episode, 2056 charging or discharging activities were recorded for the agent with access to forecasts, compared to only 1179 for the agent without forecasts. PPO charged or discharged the battery 1175 times without forecasts and only 1491 times when provided with forecasts. This reveals that the performance difference between the two DRL models is due to the greater increase in battery cycles for DQN. The increase in activity does shorten battery life, but it is factored in the reward function by increased degradation cost (see Eq. \ref{eq4}). Furthermore, after a discharge, the agent with forecasts tended to wait longer before recharging and thereby purchased electricity at lower prices. This behavior is well visible around February 19th in Fig. \ref{fig:sample}. Similarly, the agent with forecasts exploited high electricity prices during price spikes better and discharged the battery mostly at the peak. The agent without forecasts discharged the battery a few hours earlier when prices were increasing but still lower, for example on February 16th. For the entire episode, access to forecasts increased sales by CA\$350,000 whereas purchases increased by only CA\$105,000.

\begin{figure}[ht]
	\begin{center}
		\includegraphics[width=\linewidth]{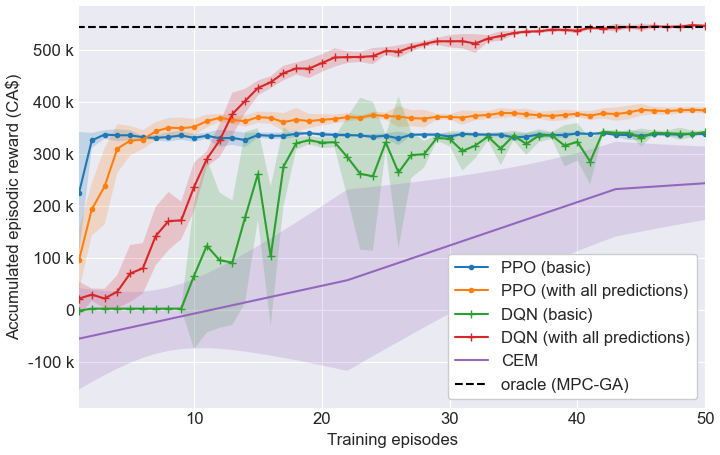}    
		\caption{Training progress of tested models with and without forecasts. All results were averaged over five independent runs. The shaded areas show $\pm$ one standard deviation.}
		\label{fig:rl_training}
	\end{center}
\end{figure}

Only the combination of forecasters for multiple horizons significantly improved results. We assume that this is due to a “majority vote” mechanism, in which the presence of multiple forecasts helps the agent to recognize the right tendency. With fewer forecasts available, due to high forecasting errors, the “majority vote” becomes less accurate leading to suboptimal policies. Figure \ref{fig:rl_training} shows the training progress of DQN and PPO in the basic experiments and with all forecasts. All RL experiments outperformed the CEM. While due to a different exploration behavior PPO is more sample efficient than DQN, it did not improve much with access to forecasts. On the other side, DQN benefited from forecasts not only regarding rewards but also in terms of stability. A possible explanation for the poor performance of PPO is that one of its key strengths, the fine-grained control of continuous action spaces, is not required for price-based arbitrage but makes learning a policy more difficult. The simpler DQN, assigning values to state-action pairs, might be better suited to recognize long-term dependencies by utilizing predictions. As charging the battery is linked with negative rewards, and the corresponding profits can occur many time-steps later, this characteristic might explain the success of DQN. 

Finally, it is important to note that the oracle does not provide an optimal solution and that its performance highly depends on the allocated computational resources. Conducting repeated optimizations over receding horizons, the quality of the solution depends on parameters such as the population size and number of iterations of the genetic algorithm. Table \ref{tbl:compute_time} compares the compute times and rewards of the models from Fig. \ref{fig:rl_training} on the same system (64-bit Windows 11 Pro, Intel Core i9-12900K CPU, 64GB RAM, and NVIDIA GeForce RTX 3080 GPU). The MPC-GA algorithm in this case study required almost 12 hours of runtime, while the best performing DQN with price predictions completed training after 2 hours on the same system. Table \ref{tbl:compute_time} further shows that the performance gains from adding predictions come with higher compute cost due to larger state spaces and repeated querying of the pre-trained predictors during DRL training.

\begin{table}[ht]
	\begin{center}
		\caption{Average compute times and rewards of the models shown in Fig. \ref{fig:rl_training}}
		\label{tbl:compute_time}
		\begin{tabular}{l|l|l}
                \textbf{Model} & \textbf{\makecell[l]{Compute \\time (s)}} & \textbf{\makecell[l]{Acc. episodic\\reward (CA\$)}} \\ \hline
                PPO (basic)& 603 & 339k \\
                PPO (with predictions) & 4,620 & 384k \\
                DQN (basic) & 1,529 & 341k \\
                DQN (with predictions) & 7,341 & 547k \\
                CEM & 2,490 & 241k \\
                Oracle (MPC-GA) & 42,210 & 546k \\
		\end{tabular}
        \vspace{-0.3cm}  
	\end{center}
\end{table}


\section{Conclusions} \label{sec:6_conclusions}

This study showed that time-series forecasting can substantially improve the performance of RL on price-based arbitrage with batteries. Despite high forecasting errors, rewards increased by 60\% in our case study when DQN was supplied with predictions on future electricity prices. The key takeaway from our experiments is that these performance gains are only possible if multiple forecasts for different horizons are combined.  Future work will have to investigate if our findings generalize to other locations and battery dispatch tasks with more uncertain variables. Besides, updating the weights of the forecasters with unseen data during agent-environment interaction could further improve results in the future.


\section*{Acknowledgments}
This work was supported by the McGill Engineering Doctoral Award (MEDA).



\bibliographystyle{asmeconf}  
\bibliography{references}

\begin{thebibliography}{10}
\newcommand{\enquote}[1]{``#1''}
\providecommand{\url}[1]{\texttt{#1}}
\providecommand{\urlprefix}{URL }
\expandafter\ifx\csname urlstyle\endcsname\relax
  \providecommand{\doi}[1]{DOI \discretionary{}{}{}#1}\else
  \providecommand{\doi}{DOI \discretionary{}{}{}\begingroup
  \urlstyle{rm}\Url}\fi
\providecommand{\eprint}[2][]{\urlprefix\url{#1#2}}

\bibitem{a213_CAO_energyarbitrage}
Cao, Jun, Harrold, Dan, Fan, Zhong, Morstyn, Thomas, Healey, David and Li,
  Kang.
\newblock \enquote{Deep Reinforcement Learning-Based Energy Storage Arbitrage
  With Accurate Lithium-Ion Battery Degradation Model.}
\newblock \textit{IEEE Transactions on Smart Grid} Vol.~11 No.~5 (2020): pp.
  4513--4521.
\newblock \doi{10.1109/TSG.2020.2986333}.

\bibitem{a241_wang_ea_with_rl}
Wang, Hao and Zhang, Baosen.
\newblock \enquote{Energy Storage Arbitrage in Real-Time Markets via
  Reinforcement Learning.}
\newblock \textit{2018 IEEE Power \& Energy Society General Meeting (PESGM)}:
  pp. 1--5. 2018.
\newblock \doi{10.1109/PESGM.2018.8586321}.

\bibitem{volatility2_RINTAMAKI2017270}
Rintamäki, Tuomas, Siddiqui, Afzal~S. and Salo, Ahti.
\newblock \enquote{Does renewable energy generation decrease the volatility of
  electricity prices? An analysis of Denmark and Germany.}
\newblock \textit{Energy Economics} Vol.~62 (2017): pp. 270--282.
\newblock \doi{10.1016/j.eneco.2016.12.019}.

\bibitem{volatility3_CIARRETA2020104749}
Ciarreta, Aitor, Pizarro-Irizar, Cristina and Zarraga, Ainhoa.
\newblock \enquote{Renewable energy regulation and structural breaks: An
  empirical analysis of Spanish electricity price volatility.}
\newblock \textit{Energy Economics} Vol.~88 (2020): p. 104749.
\newblock \doi{10.1016/j.eneco.2020.104749}.

\bibitem{x2_BIGGINS2022104234}
Biggins, F.A.V., Homan, S., Ejeh, J.O. and Brown, S.
\newblock \enquote{To trade or not to trade: Simultaneously optimising battery
  storage for arbitrage and ancillary services.}
\newblock \textit{Journal of Energy Storage} Vol.~50 (2022): p. 104234.
\newblock \doi{10.1016/j.est.2022.104234}.

\bibitem{x3_METZ201827}
Metz, Dennis and Saraiva, João~Tomé.
\newblock \enquote{Use of battery storage systems for price arbitrage
  operations in the 15- and 60-min German intraday markets.}
\newblock \textit{Electric Power Systems Research} Vol. 160 (2018): pp. 27--36.
\newblock \doi{10.1016/j.epsr.2018.01.020}.

\bibitem{a176_zhang2020}
Zhang, Zidong, Zhang, Dongxia and Qiu, Robert~C.
\newblock \enquote{Deep reinforcement learning for power system applications:
  An overview.}
\newblock \textit{CSEE Journal of Power and Energy Systems} Vol.~6 No.~1
  (2020): pp. 213--225.
\newblock \doi{10.17775/CSEEJPES.2019.00920}.

\bibitem{a171_MENG202113}
Meng, Fanyi, Bai, Yang and Jin, Jingliang.
\newblock \enquote{An advanced real-time dispatching strategy for a distributed
  energy system based on the reinforcement learning algorithm.}
\newblock \textit{Renewable Energy} Vol. 178 (2021): pp. 13--24.
\newblock \doi{10.1016/j.renene.2021.06.032}.

\bibitem{h1_9334858}
Lujano-Rojas, Juan~M., Yusta, José~M., Domínguez-Navarro, José~A., Osório,
  Gerardo~J., Shafie-khah, Miadreza, Wang, Fei and Catalão, João P.~S.
\newblock \enquote{Combining Genetic and Gravitational Search Algorithms for
  the Optimal Management of Battery Energy Storage Systems in Real-Time Pricing
  Markets.}
\newblock \textit{2020 IEEE Industry Applications Society Annual Meeting}: pp.
  1--7. 2020.
\newblock \doi{10.1109/IAS44978.2020.9334858}.

\bibitem{h2_FENG2022105508}
Feng, Lu, Zhang, Xinjing, Li, Chengyuan, Li, Xiaoyu, Li, Bin, Ding, Jie, Zhang,
  Chao, Qiu, Han, Xu, Yujie and Chen, Haisheng.
\newblock \enquote{Optimization analysis of energy storage application based on
  electricity price arbitrage and ancillary services.}
\newblock \textit{Journal of Energy Storage} Vol.~55 (2022): p. 105508.
\newblock \doi{10.1016/j.est.2022.105508}.

\bibitem{a93_zhang2019deep}
Zhang, Bin, Hu, Weihao, Cao, Di, Huang, Qi, Chen, Zhe and Blaabjerg, Frede.
\newblock \enquote{Deep reinforcement learning--based approach for optimizing
  energy conversion in integrated electrical and heating system with renewable
  energy.}
\newblock \textit{Energy Conversion and Management} Vol. 202 (2019): p. 112199.
\newblock \doi{10.1016/j.enconman.2019.112199}.

\bibitem{a79_mnih2015human}
Mnih, Volodymyr, Kavukcuoglu, Koray, Silver, David, Rusu, Andrei~A, Veness,
  Joel, Bellemare, Marc~G, Graves, Alex, Riedmiller, Martin, Fidjeland,
  Andreas~K, Ostrovski, Georg et~al.
\newblock \enquote{Human-level control through deep reinforcement learning.}
\newblock \textit{nature} Vol. 518 No. 7540 (2015): pp. 529--533.
\newblock \doi{10.1038/nature14236}.

\bibitem{a121_schulmanPPO}
Schulman, John, Wolski, Filip, Dhariwal, Prafulla, Radford, Alec and Klimov,
  Oleg.
\newblock \enquote{Proximal policy optimization algorithms.}
\newblock \textit{arXiv preprint arXiv:1707.06347}  (2017).

\bibitem{j3_arima_price_forecasting_and_feature_selection_8665889}
Bissing, Daniel, Klein, Michael~T., Chinnathambi, Radhakrishnan~Angamuthu,
  Selvaraj, Daisy~Flora and Ranganathan, Prakash.
\newblock \enquote{A Hybrid Regression Model for Day-Ahead Energy Price
  Forecasting.}
\newblock \textit{IEEE Access} Vol.~7 (2019): pp. 36833--36842.
\newblock \doi{10.1109/ACCESS.2019.2904432}.

\bibitem{j7_microeconomic_AB_https://doi.org/10.1002/asmb.2681}
Castañeda-Leyva, Netzahualcóyotl, Hernández-Ramos, Hugo, Pérez-Hernández,
  Leonel~Ramón and Rodríguez-Narciso, Silvia.
\newblock \enquote{Prediction of electricity prices for non-regulated markets
  based on a power transformed mean reverting process.}
\newblock \textit{Applied Stochastic Models in Business and Industry} Vol.~38
  No.~4 (2022): pp. 677--694.
\newblock \doi{10.1002/asmb.2681}.

\bibitem{j4_lstm_pv8483960}
He, Hui, Hu, Ran, Zhang, Yaning, Zhang, Ying and Jiao, Runhai.
\newblock \enquote{A Power Forecasting Approach for PV Plant based on
  Irradiance Index and LSTM.}
\newblock \textit{2018 37th Chinese Control Conference (CCC)}: pp. 9404--9409.
  2018.
\newblock \doi{10.23919/ChiCC.2018.8483960}.

\bibitem{j5_cnn_lstm_hybrid_comparison_DEHGHANI2023102119}
Dehghani, Adnan, Moazam, Hamza Mohammad Zakir~Hiyat, Mortazavizadeh,
  Fatemehsadat, Ranjbar, Vahid, Mirzaei, Majid, Mortezavi, Saber, Ng, Jing~Lin
  and Dehghani, Amin.
\newblock \enquote{Comparative evaluation of LSTM, CNN, and ConvLSTM for hourly
  short-term streamflow forecasting using deep learning approaches.}
\newblock \textit{Ecological Informatics} Vol.~75 (2023): p. 102119.
\newblock \doi{10.1016/j.ecoinf.2023.102119}.

\bibitem{j6_lstmcnnhybrid_implementation_9356582}
Farsi, Behnam, Amayri, Manar, Bouguila, Nizar and Eicker, Ursula.
\newblock \enquote{On Short-Term Load Forecasting Using Machine Learning
  Techniques and a Novel Parallel Deep LSTM-CNN Approach.}
\newblock \textit{IEEE Access} Vol.~9 (2021): pp. 31191--31212.
\newblock \doi{10.1109/ACCESS.2021.3060290}.

\bibitem{j8_lstm_successPENG20181301}
Peng, Lu, Liu, Shan, Liu, Rui and Wang, Lin.
\newblock \enquote{Effective long short-term memory with differential evolution
  algorithm for electricity price prediction.}
\newblock \textit{Energy} Vol. 162 (2018): pp. 1301--1314.
\newblock \doi{10.1016/j.energy.2018.05.052}.

\bibitem{j9_attention_price_lstmMENG2022124212}
Meng, Anbo, Wang, Peng, Zhai, Guangsong, Zeng, Cong, Chen, Shun, Yang, Xiaoyi
  and Yin, Hao.
\newblock \enquote{Electricity price forecasting with high penetration of
  renewable energy using attention-based LSTM network trained by crisscross
  optimization.}
\newblock \textit{Energy} Vol. 254 (2022): p. 124212.
\newblock \doi{10.1016/j.energy.2022.124212}.

\bibitem{j10_attention_hybrid_PV_paperQU2021120996}
Qu, Jiaqi, Qian, Zheng and Pei, Yan.
\newblock \enquote{Day-ahead hourly photovoltaic power forecasting using
  attention-based CNN-LSTM neural network embedded with multiple relevant and
  target variables prediction pattern.}
\newblock \textit{Energy} Vol. 232 (2021): p. 120996.
\newblock \doi{10.1016/j.energy.2021.120996}.

\bibitem{j11_smoothing_livieris_smoothing_2021}
Livieris, Ioannis~E., Stavroyiannis, Stavros, Iliadis, Lazaros and Pintelas,
  Panagiotis.
\newblock \enquote{Smoothing and stationarity enforcement framework for deep
  learning time-series forecasting.}
\newblock \textit{Neural Computing and Applications} Vol.~33 No.~20 (2021): pp.
  14021--14035.
\newblock \doi{10.1007/s00521-021-06043-1}.

\bibitem{a242_han_rl_and_dl_for_ea}
Han, Gwangwoo, Lee, Sanghun, Lee, Jaemyung, Lee, Kangyong and Bae, Joongmyeon.
\newblock \enquote{Deep-learning- and reinforcement-learning-based profitable
  strategy of a grid-level energy storage system for the smart grid.}
\newblock \textit{Journal of Energy Storage} Vol.~41 (2021): p. 102868.
\newblock \doi{10.1016/j.est.2021.102868}.

\bibitem{a225_huang}
Huang, Bin and Wang, Jianhui.
\newblock \enquote{Deep-Reinforcement-Learning-Based Capacity Scheduling for
  PV-Battery Storage System.}
\newblock \textit{IEEE Transactions on Smart Grid} Vol.~12 No.~3 (2021): pp.
  2272--2283.
\newblock \doi{10.1109/TSG.2020.3047890}.

\bibitem{a237_HARROLD2022121958}
Harrold, Daniel~J.B., Cao, Jun and Fan, Zhong.
\newblock \enquote{Data-driven battery operation for energy arbitrage using
  rainbow deep reinforcement learning.}
\newblock \textit{Energy} Vol. 238 (2022): p. 121958.
\newblock \doi{10.1016/j.energy.2021.121958}.

\bibitem{a193_DASILVAANDRE2022108551}
{da Silva André}, Joel, Stai, Eleni, Stanojev, Ognjen and Hug, Gabriela.
\newblock \enquote{Battery control with lookahead constraints in distribution
  grids using reinforcement learning.}
\newblock \textit{Electric Power Systems Research} Vol. 211 (2022): p. 108551.
\newblock \doi{10.1016/j.epsr.2022.108551}.

\bibitem{a226_DONG2021107229}
Dong, Yi, Dong, Zhen, Zhao, Tianqiao and Ding, Zhengtao.
\newblock \enquote{A Strategic Day-ahead bidding strategy and operation for
  battery energy storage system by reinforcement learning.}
\newblock \textit{Electric Power Systems Research} Vol. 196 (2021): p. 107229.
\newblock \doi{10.1016/j.epsr.2021.107229}.

\bibitem{sage2024}
Sage, Manuel and Zhao, Yaoyao~Fiona.
\newblock \enquote{Deep Reinforcement Learning for Economic Battery Dispatch: A
  Comprehensive Comparison of Algorithms and Experiment Design Choices.}
  (2024).
\newblock Manuscript under review.

\bibitem{a66_sutton2018reinforcement}
Sutton, Richard~S and Barto, Andrew~G.
\newblock \textit{Reinforcement learning: An introduction}.
\newblock MIT press (2018).

\bibitem{sage2023battery}
Sage, Manuel, Staniszewski, Martin and Zhao, Yaoyao~Fiona.
\newblock \enquote{Economic Battery Storage Dispatch with Deep Reinforcement
  Learning from Rule-Based Demonstrations.}
\newblock \textit{2023 International Conference on Control, Automation and
  Diagnosis (ICCAD)}: pp. 1--6. 2023.
\newblock \doi{10.1109/ICCAD57653.2023.10152299}.

\bibitem{a194_subramanya9777914}
Subramanya, Rakshith, Sierla, Seppo~A. and Vyatkin, Valeriy.
\newblock \enquote{Exploiting Battery Storages With Reinforcement Learning: A
  Review for Energy Professionals.}
\newblock \textit{IEEE Access} Vol.~10 (2022): pp. 54484--54506.
\newblock \doi{10.1109/ACCESS.2022.3176446}.

\bibitem{original_CNN}
Lecun, Y., Bottou, L., Bengio, Y. and Haffner, P.
\newblock \enquote{Gradient-based learning applied to document recognition.}
\newblock \textit{Proceedings of the IEEE} Vol.~86 No.~11 (1998): pp.
  2278--2324.
\newblock \doi{10.1109/5.726791}.

\bibitem{j12_Original_LSTM}
Hochreiter, Sepp and Schmidhuber, Jürgen.
\newblock \enquote{Long Short-term Memory.}
\newblock \textit{Neural computation} Vol.~9 (1997): pp. 1735--80.
\newblock \doi{10.1162/neco.1997.9.8.1735}.

\bibitem{vaswani2017attention}
Vaswani, Ashish, Shazeer, Noam, Parmar, Niki, Uszkoreit, Jakob, Jones, Llion,
  Gomez, Aidan~N, Kaiser, {\L}ukasz and Polosukhin, Illia.
\newblock \enquote{Attention is all you need.}
\newblock \textit{Advances in neural information processing systems} Vol.~30
  (2017).

\bibitem{a229_tran_bes}
Tran, Duong and Khambadkone, Ashwin~M.
\newblock \enquote{Energy Management for Lifetime Extension of Energy Storage
  System in Micro-Grid Applications.}
\newblock \textit{IEEE Transactions on Smart Grid} Vol.~4 No.~3 (2013): pp.
  1289--1296.
\newblock \doi{10.1109/TSG.2013.2272835}.

\bibitem{a230_cheng_bes}
Cheng, Yu-Shan, Liu, Yi-Hua, Hesse, Holger~C., Naumann, Maik, Truong, Cong~Nam
  and Jossen, Andreas.
\newblock \enquote{A PSO-Optimized Fuzzy Logic Control-Based Charging Method
  for Individual Household Battery Storage Systems within a Community.}
\newblock \textit{Energies} Vol.~11 No.~2 (2018).
\newblock \doi{10.3390/en11020469}.

\bibitem{w23_cole_cost}
Cole, Wesley and Karmakar, Akash.
\newblock \enquote{Cost Projections for Utility-Scale Battery Storage: 2023
  Update.}
\newblock \textit{National Renewable Energy Lab (NREL), Golden, CO (United
  States)}  (2023)\doi{10.2172/1984976}.

\bibitem{aeso}
{Alberta Electric System Operator}.
\newblock \enquote{Market and system reporting.} (2023).
\newblock \url{https://www.aeso.ca/market/market-and-system-reporting/},
  visisted on 2023-05-27.

\bibitem{era5}
Hersbach, Hans, Bell, Bill, Berrisford, Paul, Hirahara, Shoji, Hor{\'a}nyi,
  Andr{\'a}s, Mu{\~n}oz-Sabater, Joaqu{\'\i}n, Nicolas, Julien, Peubey, Carole,
  Radu, Raluca, Schepers, Dinand et~al.
\newblock \enquote{The {ERA5} global reanalysis.}
\newblock \textit{Quarterly Journal of the Royal Meteorological Society} Vol.
  146 No. 730 (2020): pp. 1999--2049.
\newblock \doi{10.1002/qj.3803}.

\bibitem{cem}
Szita, István and Lörincz, András.
\newblock \enquote{Learning Tetris Using the Noisy Cross-Entropy Method.}
\newblock \textit{Neural Computation} Vol.~18 No.~12 (2006): pp. 2936--2941.
\newblock \doi{10.1162/neco.2006.18.12.2936}.

\bibitem{sincos2_app12199788}
Huang, Junhui, Algahtani, Mohammed and Kaewunruen, Sakdirat.
\newblock \enquote{Energy Forecasting in a Public Building: A Benchmarking
  Analysis on Long Short-Term Memory (LSTM), Support Vector Regression (SVR),
  and Extreme Gradient Boosting (XGBoost) Networks.}
\newblock \textit{Applied Sciences} Vol.~12 No.~19 (2022).
\newblock \doi{10.3390/app12199788}.

\end{thebibliography}

\end{document}